\documentclass[10pt,twocolumn,letterpaper]{article}

\usepackage[pagenumbers]{cvpr} 

\usepackage{bm}
\usepackage{graphicx}
\usepackage{amsmath}
\usepackage{amssymb}
\usepackage{booktabs}
\usepackage{booktabs}
\usepackage{multirow}
\usepackage[table]{xcolor}
\definecolor{lightgray}{gray}{0.9}
\usepackage[pagebackref,breaklinks,colorlinks]{hyperref}

\usepackage[capitalize]{cleveref}
\crefname{section}{Sec.}{Secs.}
\Crefname{section}{Section}{Sections}
\Crefname{table}{Table}{Tables}
\crefname{table}{Tab.}{Tabs.}

\usepackage{appendix}

\def\cvprPaperID{2630} 
\def\confName{CVPR}
\def\confYear{2022}

\def\wrt{{w}\onedot{r}\onedot{t}\onedot}
\def\etal{\emph{et~al}\onedot}

\def\forexample{\emph{e.g}\onedot}
\def\thatis{\emph{i.e}\onedot}

\def\secmk{Sec.~}
\def\figmk{Fig.~}
\def\tablemk{Tab.~}

\def\supmat{\textbf{Sup.~Mat~}}

\def\numobj{20~}
\def\numsbj{20~}
\def\numcam{25~}
\def\numscene{1K~} 
\def\VCLhoDBname{CBF~}

\begin{document}
\title{Stability-driven Contact Reconstruction From Monocular Color Images}

\author{Zimeng Zhao\hspace{2mm}\hspace{5mm} 
Binghui Zuo\hspace{2mm}\hspace{5mm} 
Wei Xie\hspace{2mm}\hspace{5mm} 
Yangang Wang\footnotemark[1]\\%
\\
Southeast University, China\\
}

\maketitle
\begin{abstract}
   Physical contact provides additional constraints for hand-object state reconstruction as well as a basis for further understanding of interaction affordances. Estimating these severely occluded regions from monocular images presents a considerable challenge. 
   Existing methods optimize the hand-object contact driven by distance threshold or prior from contact-labeled datasets. However, due to the number of subjects and objects involved in these indoor datasets being limited, the learned contact patterns could not be generalized easily. 
   Our key idea is to reconstruct the contact pattern directly from monocular images, and then utilize the physical stability criterion in the simulation to optimize it. This criterion is defined by the resultant forces and contact distribution computed by the physics engine. 
   Compared to existing solutions, our framework can be adapted to more personalized hands and diverse object shapes. Furthermore, an interaction dataset with extra physical attributes is created to verify the sim-to-real consistency of our methods. Through comprehensive evaluations, hand-object contact can be reconstructed with both accuracy and stability by the proposed framework. 
\end{abstract}

\renewcommand{\thefootnote}{\fnsymbol{footnote}}
\footnotetext[1]{Corresponding author. E-mail: yangangwang@seu.edu.cn. This work was supported in part by the National Natural Science Foundation of China (No. 62076061), the ``Young Elite Scientists Sponsorship Program by CAST" (No. YES20200025), and the ``Zhishan Young Scholar" Program of Southeast University (No. 2242021R41083).}

\section{Introduction} 
Monocular hand-object contact recovery has wide applications, which can enable accurate interactions in metaverse and telepresence robot control. 
Traditional methods often judge contact regions by the closest distances between surfaces of the hand and object in an optimization strategy~\cite{tzionas2016capturing}, where the recovered contact highly depends on the accuracy of hand-object pose estimation. 
However, this accuracy is hard to be guaranteed by monocular reconstruction. 
Recent approaches~\cite{grady2021contactopt,cao2021reconstructing, yang2021cpf} learn the hand-object contact prior from well-labeled datasets~\cite{taheri2020grab, brahmbhatt2020contactpose}, yet their performance rely on the diversity of the contact data. 
\begin{figure}[!t]
    \centering
    \includegraphics[width=\linewidth]{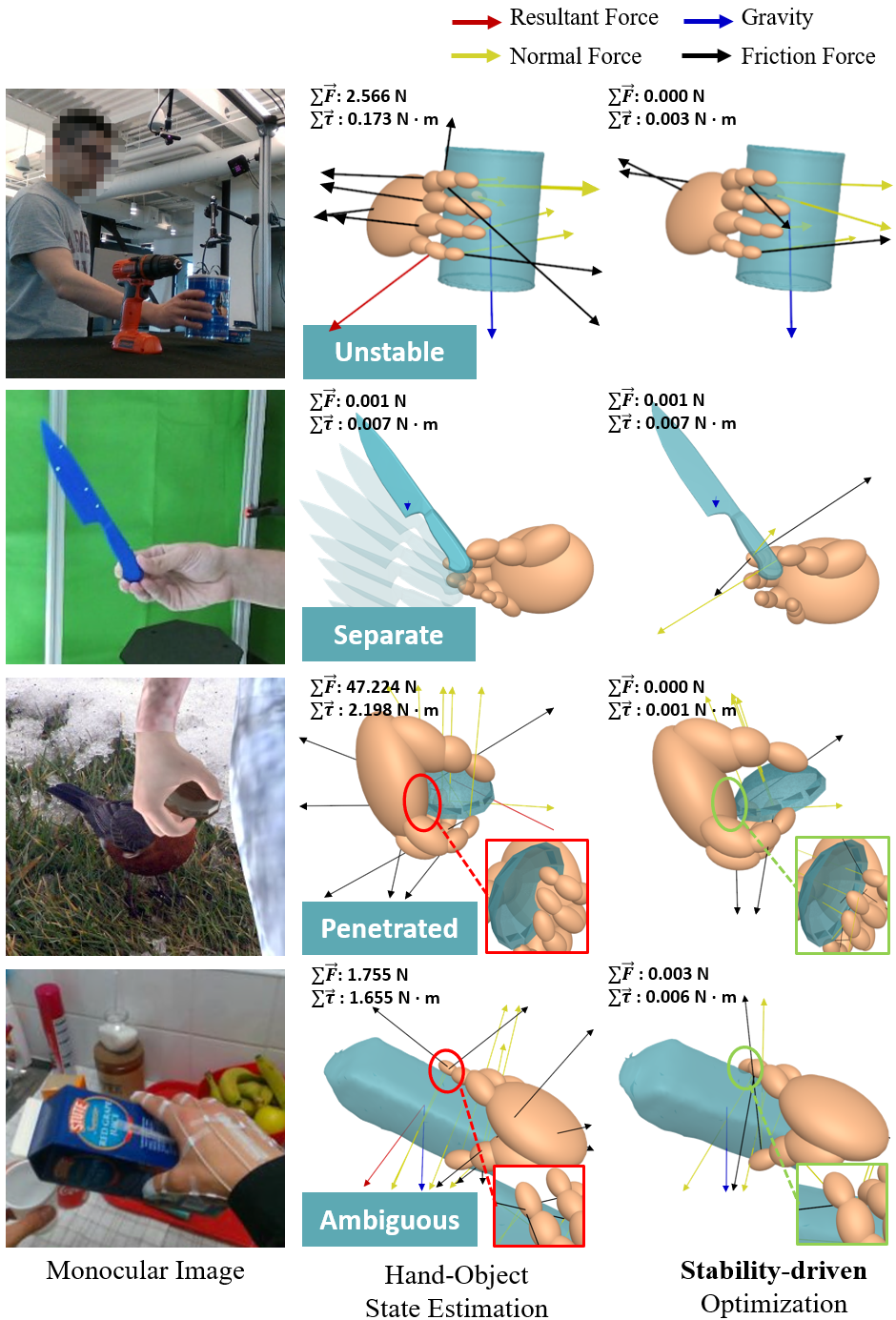}
    \caption{\textbf{Stability-driven contact reconstruction}. Each row illustrates the hand-object state represented by multiple ellipsoids. The resultant forces and torques on the objects are calculated by the physics engine~\cite{coumans2013bullet}. }
    \vspace{-4mm}
    \label{teaser_optRes}
\end{figure}

Generally speaking, a credible contact is to ensure the interacting stability between a hand and object in the physical world, either to keep the object stationary or with a required acceleration.
To reconstruct this stable contact, 
our key idea is to \textbf{reconstruct the contact pattern driven by the physical criteria (\ie, the balancing of forces and torques) calculated by a physics engine.} 
It is noted that most existing methods~\cite{hasson2019learning, corona2020ganhand, karunratanakul2020grasping, jiang2021hand, christen2021d} utilize the relative object displacements to evaluate the contact stability. However, the criteria cannot be directly used to drive optimization due to the shortcomings in both hand modeling and stability evaluation. 

Regarding hand modeling, traditional methods for simulation utilize either a whole mesh~\cite{hasson2019learning} or multiple mesh segments without connectivity~\cite{tzionas2016capturing,mamou2016volumetric,thul2018approximate}. Such models are difficult to perform robot control and force analysis due to the lack of a kinematic tree. To overwhelm the limitations, we adopt a structured multi-body for dynamics simulation, whose rigid parts can be automatically adjusted according to the personalized information estimated from an image. Specifically, those hand rigid parts and the object are jointly represented as a series of ellipsoidal primitives, and our front-end network is used to estimate the state parameters for composing these primitives. Compared with MANO~\cite{romero2017embodied} parameters, regressing this state not only brings acceleration to the calculation of self and mutual collision during network training but also facilitates the construction of our multi-body in a physics engine. 

We argue that the stability could not be fully evaluated by the displacement of the contact object, which is an average effect of the resultant force. Alternatively, some novel stability criteria are proposed with more considerations on physical factors related to contact. ~Considering the contact constraints are unilateral, we use sampling-based optimization rather than gradient-based methods to make the estimated state meet the above stability requirements. A hand-object contact dataset is further built to analyze the sim-to-real gap in our simulation. In addition to images and meshes of hands and objects, our proposed dataset includes physical properties and stability evaluations for each interaction scene. The stability of a real interaction scene is mainly evaluated by the additional balancing force that needs to be applied to the object when capturing contact by our multi-view system. 

Summarily, we make the following contributions. 
\begin{itemize}
    \item A regression-optimization framework for reconstructing hand-object contacts and physical correlation from monocular images guided by stability; 
    \vspace{-2mm}
    \item A hand-object representation and learning strategy based on ellipsoid primitives, which brings convenience to the process of both deep learning inference and physical simulation; 
    \vspace{-2mm}
    \item A hand-object interaction dataset containing physical attributes and stability metrics, which validates the sim-to-real consistency of related methods.
\end{itemize}
The dataset and codes will be publicly available at ~\url{https://www.yangangwang.com}. 
\section{Related Work} 
The reconstruction method discussed in this part mainly takes the monocular color image as input and considers the interaction between one hand and one object. 

\noindent\textbf{Hand-Object State Estimation}. 
With the rapid increase of 3D hand datasets~\cite{zimmermann2019freihand, gomez2019large, zhao2020hand, xiang2017posecnn, liu2021stereobj} and object datasets~\cite{xiang2017posecnn, hinterstoisser2012model, liu2021stereobj}, data-driven methods~\cite{zimmermann2017learning, iqbal2018hand, zhang2019end, ge20193d, zhou2020monocular, moon2020i2l, kehl2017ssd, tekin2018real, hu2019segmentation, peng2019pvnet,brahmbhatt2019contactdb,wang2020rgb2hands} become popular in the community. However, when the hand interacts with the object, the problem becomes further complicated because of severe occlusions. The representation in pioneer datasets~\cite{garcia2018first} and methods~\cite{tekin2019h, doosti2020hope} only contained hand skeletons and object bounding boxes. Subsequent work~\cite{hampali2020honnotate, chao2021dexycb} provided more fine-grained hand-object surfaces depicted by MANO parameters~\cite{romero2017embodied} and specific object categories~\cite{calli2015ycb, xiang2017posecnn}. With more synthetic data, Hasson~\etal~\cite{hasson2019learning} explored the scheme to reconstruct the shape and pose of hand-object through a unified network. Other methods~\cite{hasson2020leveraging, grady2021contactopt,cao2021reconstructing,hasson2021towards} placed more emphasis on the hand state and object pose. 
This work also relies on providing object meshes in the simulation. However, the object pose and hand features are estimated from the input images.

\noindent\textbf{Contact Estimation}. 
There is a trend~\cite{narasimhaswamy2020detecting, shan2020understanding} to understand the interaction pattern directly at the image level. Since the contact area is generally invisible in the image, more methods explore it from 3D states. To enable data-driven methods, many pioneers~\cite{pham2017hand, ehsani2020use, brahmbhatt2020contactpose} utilized expensive sensors, ingenious deployment, and manual labor to obtain actual contact information without affecting the hand-object appearance (marker-less). Others ~\cite{hasson2019learning, taheri2020grab} used the distance between the hand and the object surface in Mocap data as the criterion to annotate the contact. Benefit from these datasets, recent methods~\cite{grady2021contactopt,cao2021reconstructing, yang2021cpf} learn the contact area prior in advance and then iteratively optimize the hand-object state according to the prior. In the evaluation stage, some approaches treat the state with more contact coverage ratio as stable~\cite{grady2021contactopt}. But this may exacerbate unreasonable penetrations rather than improve the contact quality. As ~\cite{hasson2019learning} pointed out, this can be compensated for by evaluation methods based on the physical simulation~\cite{tzionas2016capturing}. With considering more contact-related physics, we create a stability criterion to effectively optimize the hand-object state without the prior dependence. 

\noindent\textbf{Hand Collision Shape}. 
Although hand meshes~\cite{romero2017embodied} are convenient for rendering, they require expensive computation for collision detection on each vertex\cite{tzionas2016capturing, hasson2019learning, moon2020deephandmesh, grady2021contactopt}. For articulated objects such as the human body and hand, collisions occur not only with other objects but also between different links of themselves. Several attempts~\cite{deng2020nasa, karunratanakul2020grasping, mihajlovic2021leap, karunratanakul2021skeleton} have been made to implicitly represent the surface with the neural occupancy function, but they are ineffective for self-intersection~\cite{mihajlovic2021leap}. By contrast, approximation of articulated objects using geometric primitives, \forexample capsules ~\cite{rehg1994visual, fleishman2015icpik}, spheres~\cite{sridhar2013interactive,sridhar2014real,qian2014realtime,sridhar2015fast, wan2019self, mueller2019real} or mixtures~\cite{oikonomidis2011efficient, oikonomidis2011full} are more intuitive to tackle both kind of intersections. \cite{tkach2016sphere, tkach2017online} presented a conversion method from implicit spheres to smooth triangular mesh. We propose a more concise scheme to represent the hand-object as a series of ellipsoids. It builds a bridge between network regression and optimization in the simulated environment. 
\section{Method}
We take two steps to reconstruct the state $\mathcal{S}$ of the hand-object and their physical contact $\mathcal{R}$ from monocular color images. First, a network is built to regress the shape and coarse pose of the hand-object represented by ellipsoidal parameters(\secmk\ref{sec31_hope}). The above parameters are applied to create dynamics scenes as the initial state of contact optimization(\secmk\ref{sec32_contactopt}). To facilitate the formulation, the tilde superscripts represent the variables regressed from the network, the hat superscripts represent the variables optimized in the simulation, and the star superscripts represent the ground truth. 

\begin{figure}[!t]
    \centering
    \includegraphics[width=\linewidth]{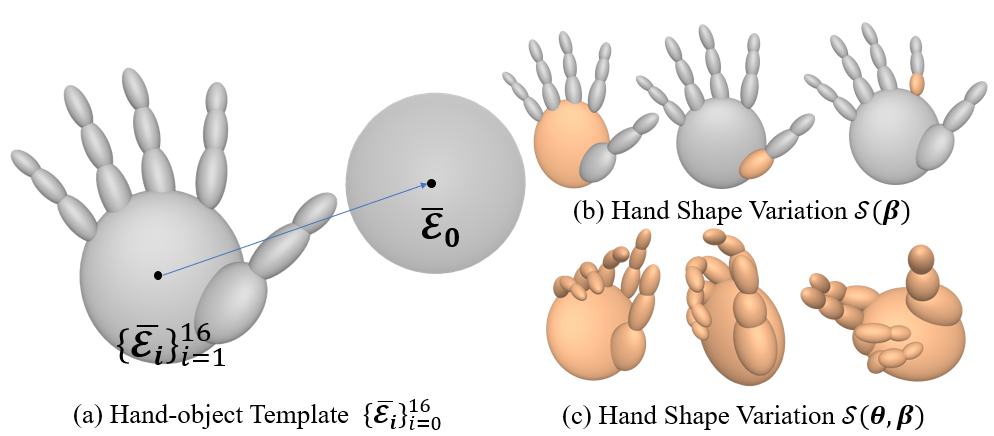}
    \caption{\textbf{Hand-object state representation}. The part colored in brown indicates a shape variation \wrt the template. (a) Hand-object template consists of 17 ellipsoids; (b) Template hands with shape variations. (c) Personalized hands with pose variations.}
    \vspace{-6mm}
    \label{fig02_elliphand}
\end{figure}

\subsection{Hand-Object State Estimation}
\label{sec31_hope}
\noindent\textbf{Ellipsoid representation}. To directly import scene into the physics engine, the states of the personalized hand and object are uniformly represented as a series of ellipsoids rather than MANO~\cite{romero2017embodied}. Specifically, a hand is approximated as 16 articulated ellipsoids and an object is approximated as one ellipsoid. Each ellipsoid can be implicitly represented as the zero isosurface of the quadratic form function: 
\begin{equation}
    \begin{aligned}
        \mathcal{E}(\bm{x} | \bm{c}, \bm{r}, \bm{a}) = (\bm{x} - \bm{c})^T A(\bm{r}, \bm{a}) (\bm{x} - \bm{c}) - 1
    \end{aligned}
    \label{eqn_ellipsoid}
\end{equation}
where $\bm{c}$ is the ellipsoid center, $\bm{r}$ is the radii, $\bm{a}$ is the orientation represented as axis-angle. It should be noted that the decomposition of the symmetric matrix $A(\bm{r}, \bm{a}) = R(\bm{a})^T \mathrm{diag}(\bm{r})^{-2} R(\bm{a})$ is not unique, \forexample $
A((a, b, c)^T, (0,0,0)^T) $ vs $ A((b, a, c)^T, (0,0,0.5\pi)^T)$. Therefore, we adopt a traditional strategy~\cite{romero2017embodied} to create a hand template $\{\bar{\mathcal{E}}_i\}_{i=1}^{16}$ shown in \figmk\ref{fig02_elliphand}. With this template, our hand-object state can be formulated as: 
\begin{equation}
    \begin{aligned}
        &\mathcal{S}(\{\mathcal{E}_i\}_{i=0}^{16}) = \mathcal{S} (\bm{\beta}, \bm{\theta}, \bm{\phi} ; ~\{\bar{\mathcal{E}}_i\}_{i=1}^{16})\\
        &\bm{\beta} \triangleq \{\delta\bm{r}_i\}_{i=1}^{16}, 
        \bm{\theta} \triangleq \{ \delta\bm{a}_i\}_{i=1}^{16}, \bm{\phi} \triangleq \{\delta\bm{r}_0, \delta\bm{a}_0, \delta\bm{c}_0\} \\
    \end{aligned}
    \label{eqn_hope}
\end{equation}
In this model, each ellipsoid can be scaled by $\delta\bm{r}_i$ and rotated by $\delta\bm{a}_i$ \wrt its local frame. The center of the palm is used as the coordinate origin and the camera coordinate system is adopted in the network prediction phase. Other ellipsoid centers can be constrained adaptively according to the connection of the ellipsoid to its parent. Because interacting objects usually keep a comparable scale and orientation to the palm, $\{\delta\bm{r}_0, \delta\bm{a}_0\}$ as well as the center offset $\delta\bm{c}_0$ of the object are relative to the $\mathcal{E}_1$. 

\noindent\textbf{Mesh conversion}. 
The explicit surface mesh is acquired from implicit primitives in three steps shown in \figmk\ref{fig03_convex}(a 1-3). 
According to~\cite{bloomenthal1991convolution, tkach2016sphere}, the zero isosurface of the following function corresponds a mesh surface: 
\begin{equation}
    \begin{aligned}
        \mathcal{M}(\bm{x}) = \mathrm{min}\{\mathcal{E}_i(\bm{x} | \bm{c}_i, \bm{r}_i, \bm{a}_i)\}_{i=1}^{16}
    \end{aligned}
    \label{eqn_convex}
\end{equation}
Additional convex hull calculations will make its surface smoother. We use this approach to project the reconstructed hand model into the image to calculate the error. 
On the other hand, as shown in \figmk\ref{fig03_convex}(b 1-3), diverse LBS hand meshes~\cite{tzionas2016capturing, romero2017embodied, dkulon2019rec, zhao2020hand, moon2020deephandmesh} are first segmented according to the skinning weights. Then oriented bounding box is created for each segment, and the final ellipsoid maintains the same radii and orientation as the box. This approach is used to convert those existing mesh-labeled datasets to the ground truth of $ \bm{\beta}^\star,  \bm{\theta}^\star$ for our training process. 

\begin{figure*}[!t]
    \centering
    \includegraphics[width=\linewidth]{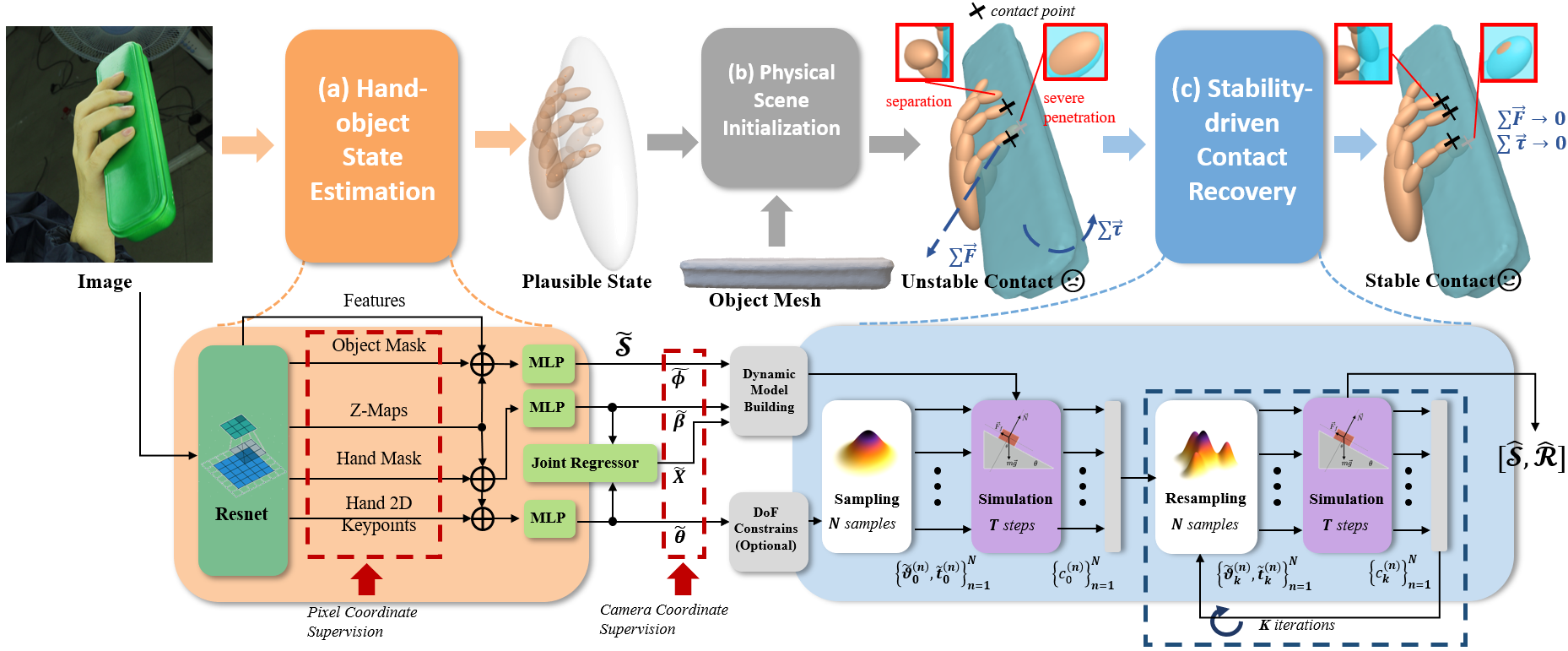}
    \caption{\textbf{Stable contact reconstruction pipeline}. (a) Hand-object state represented by implicit ellipsoids is estimated from the input image; (b) Simulated interaction scene is direct constructed from the estimated parameters; (c) The optimization process is driven by the stability cost in simulation to get more reliable states iteratively. }
    \vspace{-4mm}
    \label{fig01_pipeline}
\end{figure*}

\begin{figure}[!t]
    \centering
    \includegraphics[width=\linewidth]{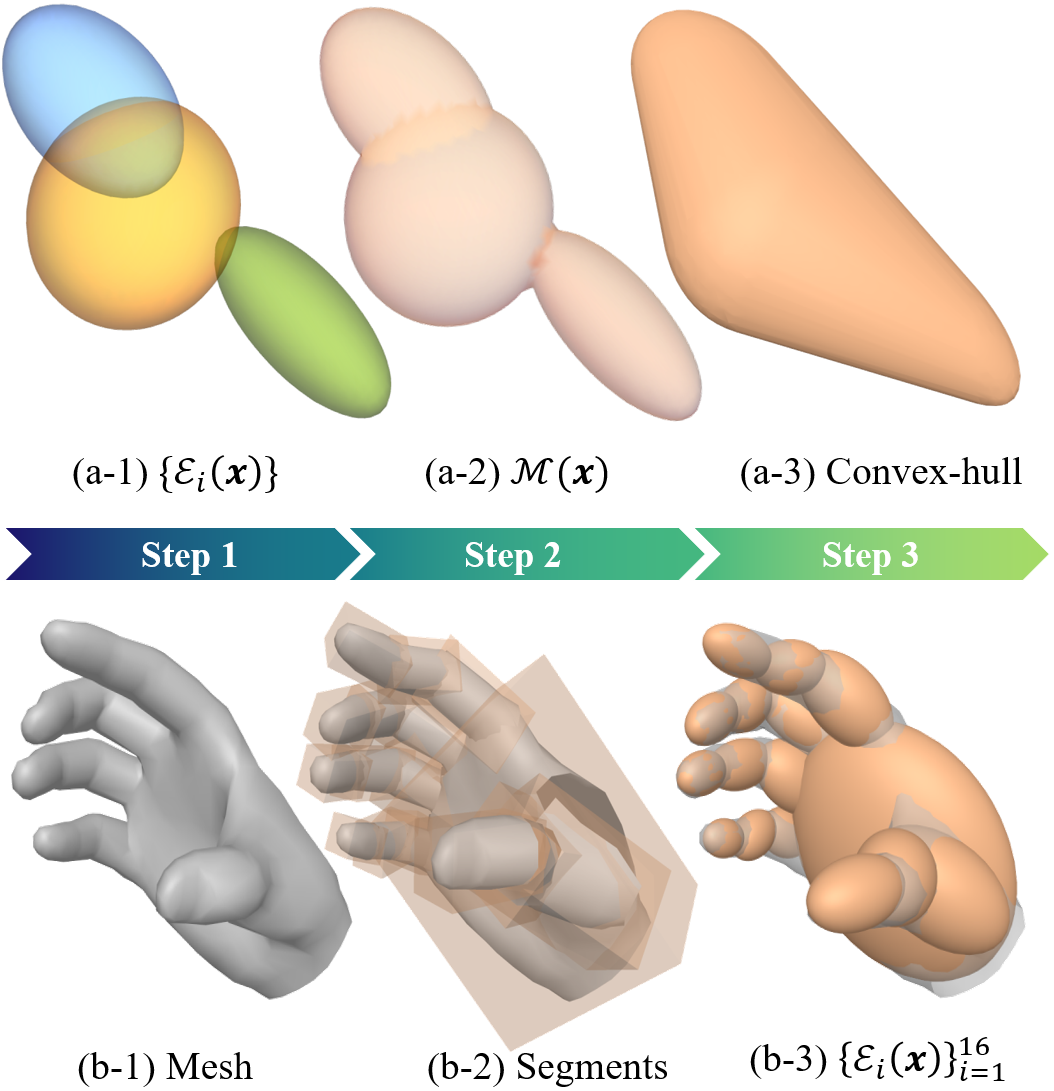}
    \caption{\textbf{Conversions between implicit and explicit hand shape}. (a 1-3) Steps from implicit ellipsoids to mesh. To show more details, 3 ellipsoids with great directional variation are used to illustrate;~ (b 1-3) Steps from explicit mesh to ellipsoids.}
    \vspace{-6mm}
    \label{fig03_convex}
\end{figure}

\noindent\textbf{Network architecture}. The network is structurally designed as an encoder-decoder. To retain more network attention on the hand-object RoI, pixel-wise features including 2D heatmap, Z-maps of hand joints and object center, hand mask, and object mask are decoded and supervised. The backbone of its encoder is ResNet18~\cite{he2016deep} with extra connections to its decoder. Those encoded features are then encoded again and concatenated with the previous features to predict our state parameters $\bm{\beta}, \bm{\theta}, \bm{\phi}$. In addition, joint regressor $\mathcal{J}(\bm{\beta}, \bm{\theta})$ is required to regress the joint position between adjacent ellipsoids. It is designed as a two-layer MLP and used to regress joint coordinates $X \in \mathbb{R}^{3\times21}$ from the explicit mesh vertices of each ellipsoid. 

\noindent\textbf{Training process}. 
Because the coordinate of our representation is hand-centered, the datasets with only hand mesh annotation~\cite{zimmermann2019freihand, dkulon2019rec, zhao2020hand, moon2020deephandmesh} can be used in our training. In the first stage, semi-supervised paradigm is adopted to pre-train our network using those datasets with hand-only or object-only annotations. The overall loss includes: 
\begin{equation}
    \begin{aligned}
        {L}_{\mathrm{S1}} &= \|\tilde{\bm{\beta}}- \bm{\beta}^\star\|_2^2 + \|\tilde{\bm{\theta}}- \bm{\theta}^\star\|_2^2 + \|\tilde{X}- X^\star\|_2^2 \\ &+ {L}_{2D} + {L}_{in}(\tilde{\mathcal{S}})
    \end{aligned}
    \label{eqn_Ls1}
\end{equation}
The first three items are the hand 3D reconstruction errors. The joint location is estimated with the help of our joint regressor $\tilde{X} = \mathcal{J}(\tilde{\bm{\beta}}, \tilde{\bm{\theta}})$. ${L}_{2D}$ contains the error of all the 2D information regressed in the intermediate steps. Some datasets may not have all annotations, then the corresponding term is also not supervised. The last term is the contact loss designed as point-based~\cite{mihajlovic2021leap, karunratanakul2021skeleton} to penalize the collision among ellipsoids: 
\begin{equation}
    \begin{aligned}
        {L}_{in}(\tilde{\mathcal{S}}) = -\sum_{\bm{x} \in \Omega(\mathcal{E}_i)} \sum_{j\neq i} \mathcal{E}_j(\bm{x} | \tilde{\mathcal{S}}), ~\mathrm{where}~\mathcal{E}_j(\cdot) < 0
    \end{aligned}
    \label{eqn_Lin}
\end{equation}
In practice, 872 vertices uniformly distributed on  $\Omega(\mathcal{E})$ are sampled in advance, whose actual coordinates $\bm{x}$ on $\mathcal{E}_i$ are determined by the ellipsoidal parameters $\tilde{\mathcal{S}}$.

In the second stage, we use the datasets with full annotations~\cite{hampali2020honnotate, hasson2019learning, brahmbhatt2020contactpose, chao2021dexycb} to train our network thoroughly: 
\begin{equation}
    \begin{aligned}
        {L}_{\mathrm{S2}} =  {L}_{\mathrm{S1}} + \|\tilde{\bm{\phi}}- \bm{\phi}^\star\|_2^2 + \|\Pi(\tilde{\mathcal{S}})- \Pi(\mathcal{S}^\star)\|_2^2
    \end{aligned}
    \label{eqn_Ls2}
\end{equation}
where $\Pi$ denotes the differentiable projection process to generate the hand and object mask through orthogonal projection. The camera parameters can be obtained by comparing the scale and translation of the hand model with 2D key-points in the image. 

$\mathcal{J}(\bm{\beta}, \bm{\theta})$ is trained independently. Since it is a mapping from the surface vertices and joints of the hand model during the movement, we obtain a large amount of pairwise training data through the forward dynamic of our hand model in the physics engine. 

\noindent\textbf{Implementation details}. Our networks are trained on a single NVIDIA GeForce RTX 3090 GPU at a base learning rate of 1e-4, an input image size of $256\times256$, and a batch size of 64, respectively. We use Adam solver~\cite{kingma2014adam} in PyTorch as the optimizer in our training. 

\begin{figure}[!t]
    \centering
    \includegraphics[width=\linewidth]{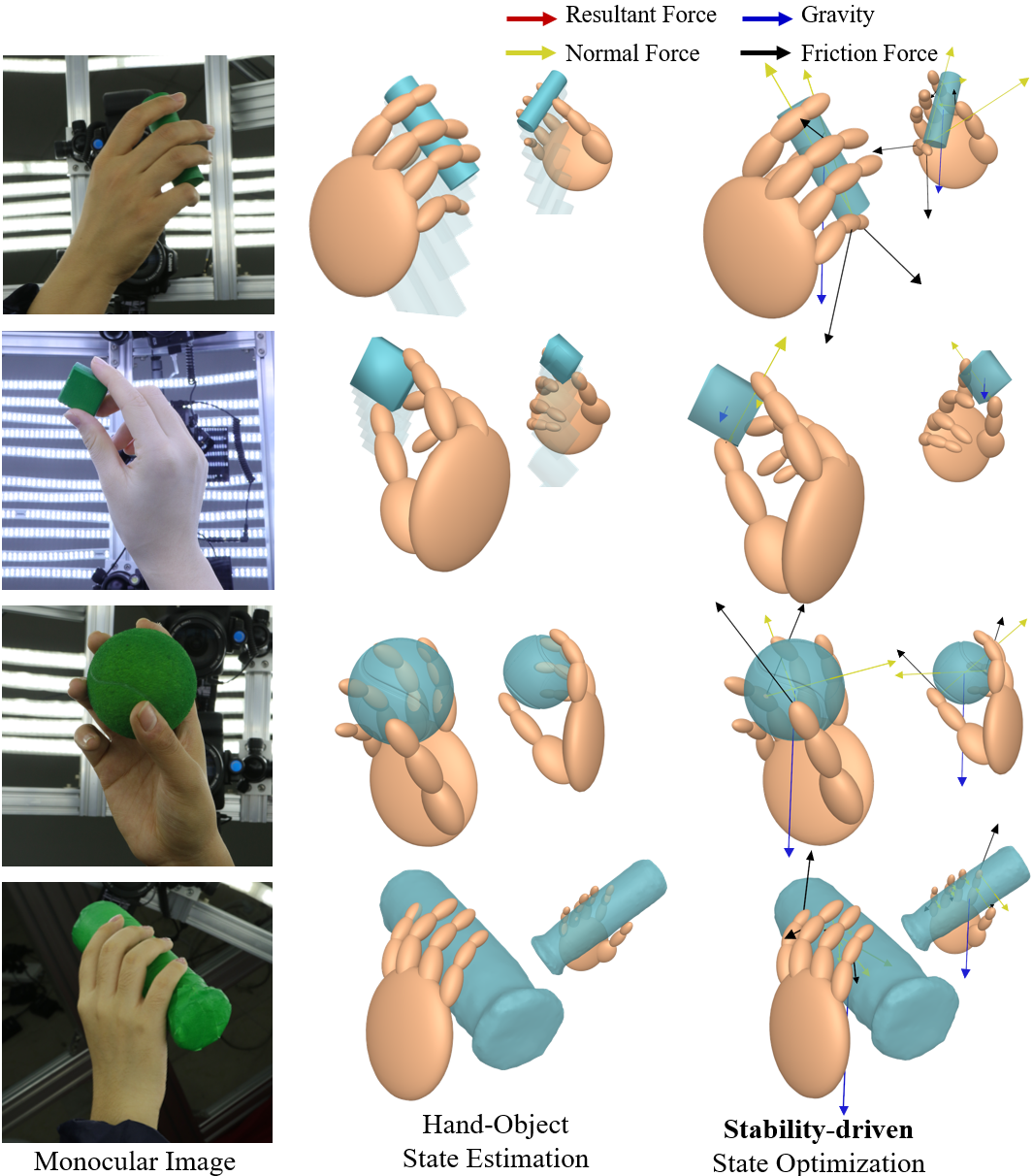}
    \caption{\textbf{Qualitative results on our dataset}. For each sample image, the estimated results from the network and optimized results are displayed from two views.}
    \vspace{-4mm}
    \label{fig04_resshow2}
\end{figure}

\subsection{Physical Contact Recovery} 
\label{sec32_contactopt}
Our optimization process is driven by the physical stability evaluated on each sample. 

\noindent\textbf{Scene initialization}. The estimated state $\tilde{\mathcal{S}}$ are used to initialize the interaction scene in the physics engine\cite{coumans2013bullet}. Firstly, the hand template $\{\bar{\mathcal{E}}_i\}_{i=1}^{16}$ with personalized variation $\tilde{\bm{\beta}}$ are used to construct a dynamic multi-body with 16 ellipsoidal links and fixed root at the origin. The detailed object mesh is loaded to the scene with the position $\tilde{\bm{p}}$ and orientation $\tilde{\bm{q}}$ which are determined by $\tilde{\bm{\phi}}$. Because it is challenging to estimate the linear acceleration $\vec{\bm{a}}$ and angular acceleration $\vec{\bm{\alpha}}$ of the object from a single image, they are simply set to zero in the following steps. To facilitate the sampling, the hand pose $\bm{\theta}$ represented by the axis-angles is converted into $\bm{\vartheta}$ represented by the Euler angles. Two schemes are adopted for local DoF: retain all 45 or only 20 physically plausible ones\cite{yang2021cpf, zhao2021travelnet}, \thatis $|\bm{\vartheta}|= {48}$ or ${23}$. The hand root is constrained at the origin before optimization, and is allowed to reach a new location $\tilde{\bm{t}}$ during sampling. As a result, $(\tilde{\bm{\vartheta}}, \tilde{\bm{t}}, \tilde{\bm{p}}, \tilde{\bm{q}})$ are involved in the next step. 

\noindent\textbf{Stability evaluation}. The actual physical contact of the given hand-object state is calculated in the impulse-based simulation~\cite{mirtich1996impulse}. Specifically, the collisions are detected among the hand links and object. Based on the Coulomb friction model~\cite{featherstone2014rigid, holl2018efficient}, normal force and lateral friction forces at each contact point are calculated based on the penetration depth. The hand is maintained at a given target pose driven by the PD controller, and the object moves passively due to its own gravity and hand contact forces. Consequently, the contact is evaluated by the stability cost: 
\begin{equation}
    \begin{aligned}
        {C} = {C}_{\mathcal{S}}(\hat{\bm{p}}, \hat{\bm{q}}, \hat{\bm{\vartheta}}, \hat{\bm{t}}) + {C}_{\mathcal{R}}(\bm{\vec{f}}, \bm{\vec{\tau}}, m)
    \end{aligned}
    \label{eqn_Criterion}
\end{equation}
where ${C}_{\mathcal{S}}$ measures the change in hand-object state before and after simulation, ${C}_{\mathcal{R}}$ measures the physical relationship including the resultant force $\bm{\vec{f}}(t)$, torque $\bm{\vec{\tau}}(t)$ and the number of contact points $m(t)$ collected in $0 < t <T$: 
\begin{equation}\left\{
    \begin{aligned}
        {C}_{\mathcal{S}} &= \|\hat{\bm{p}}-\tilde{\bm{p}}\|_2 + {L}_Q (\hat{\bm{q}}^{-1}\tilde{\bm{q}}) + \frac{1}{|\bm{\vartheta}|} \|\hat{\bm{\vartheta}}-\tilde{\bm{\vartheta}}\|_1 + \|\hat{\bm{t}}-\tilde{\bm{t}}\|_2 \\
        {C}_{\mathcal{R}} &= \frac{1}{T} \sum_{t=0}^T \frac{\|\bm{\vec{f}(t)} - M_o \vec{\bm{a}}\|_2^2}{\|M_o \vec{\bm{g}}\|_2^2} + \frac{\|\bm{\vec{\tau}}(t) - I_o \bm{\vec{\alpha}}\|_2^2}{\|I_o \bm{\vec{\alpha}}\|_2^2} + e^{-m(t)}\\
    \end{aligned}\right. 
    \label{eqn_Criterion_s}
\end{equation}
In practice, the change of object direction is measured by the angular difference. The normalization of $\bm{\vec{f}}(t)$ and  $\bm{\vec{\tau}}(t)$ based on the mass $M_o$ and moment of inertia $I_o$ of the particular object could avoid the impact of the stability cost due to the variation of objects. To prevent the object from flying out of the hand operating area in each simulation step $t$, the state of the object would be reset if $\|\hat{\bm{p}}(t)-\tilde{\bm{p}}\|_2 > 0.1$ or ${L}_Q (\hat{\bm{q}}(t)^{-1}\tilde{\bm{q}}) > 0.3 \pi$. 


\noindent\textbf{Iterative sampling}. 
Due to the contact constraint being unilateral which may fail to compute the gradient, We use sampling-based optimization driven by the above stability criterion. The distribution $D(\hat{\bm{\vartheta}})$ is initialized by Gaussian with $\tilde{\bm{\vartheta}}$ as the center and $0.1\pi$ as the variance of each dimension, and the distribution $D(\hat{\bm{t}})$ is initialized by Gaussian with $\bm{0}$ as the center and $0.05$ as the variance of each dimension. In each iteration $k$, the samples $\{\tilde{\bm{\vartheta}}_k^{(n)}, \tilde{\bm{t}}_k^{(n)}\}_{n=1}^N$ with lower cost are given greater weight. Using these weighted samples, the variance of each dimension is updated before the resampling. In the last round, the lowest cost state, together with the contact point and contact force, is the result of hand-object interaction reconstruction. 

\noindent\textbf{Implementation details}. 
In our experiment, the number of sampling iterations is set to $K = 30$, and the number of samples is set to $N = 300$. For each state sample, the interaction process is performed $T = 120$ steps in the physics engine. All the samples in the same iteration are simulated in parallel. The time step follows the default 240Hz setting in the bullet physics~\cite{coumans2013bullet}, \thatis each simulation process corresponds to $0.5 \mathrm{s}$ in the real physical world. Distance is measured in meters, mass in kilogram, and force in Newton. The gravity direction is considered to be down along the Y-axis in the image coordinates. For objects from other datasets~\cite{xiang2017posecnn, hasson2019learning,brahmbhatt2020contactpose,taheri2020grab}, the mass is proportional to its volume, and the density is uniformly set to $500 \mathrm{kg/m^{3}}$. The restitution coefficient of both hand and object is set to 1.0. The friction coefficient between the hand and objects is set to 0.8. For the objects in our dataset, the mass and friction settings follow the actual measurement results contained in our \supmat. 

\begin{figure}[!t]
    \centering
    \includegraphics[width=\linewidth]{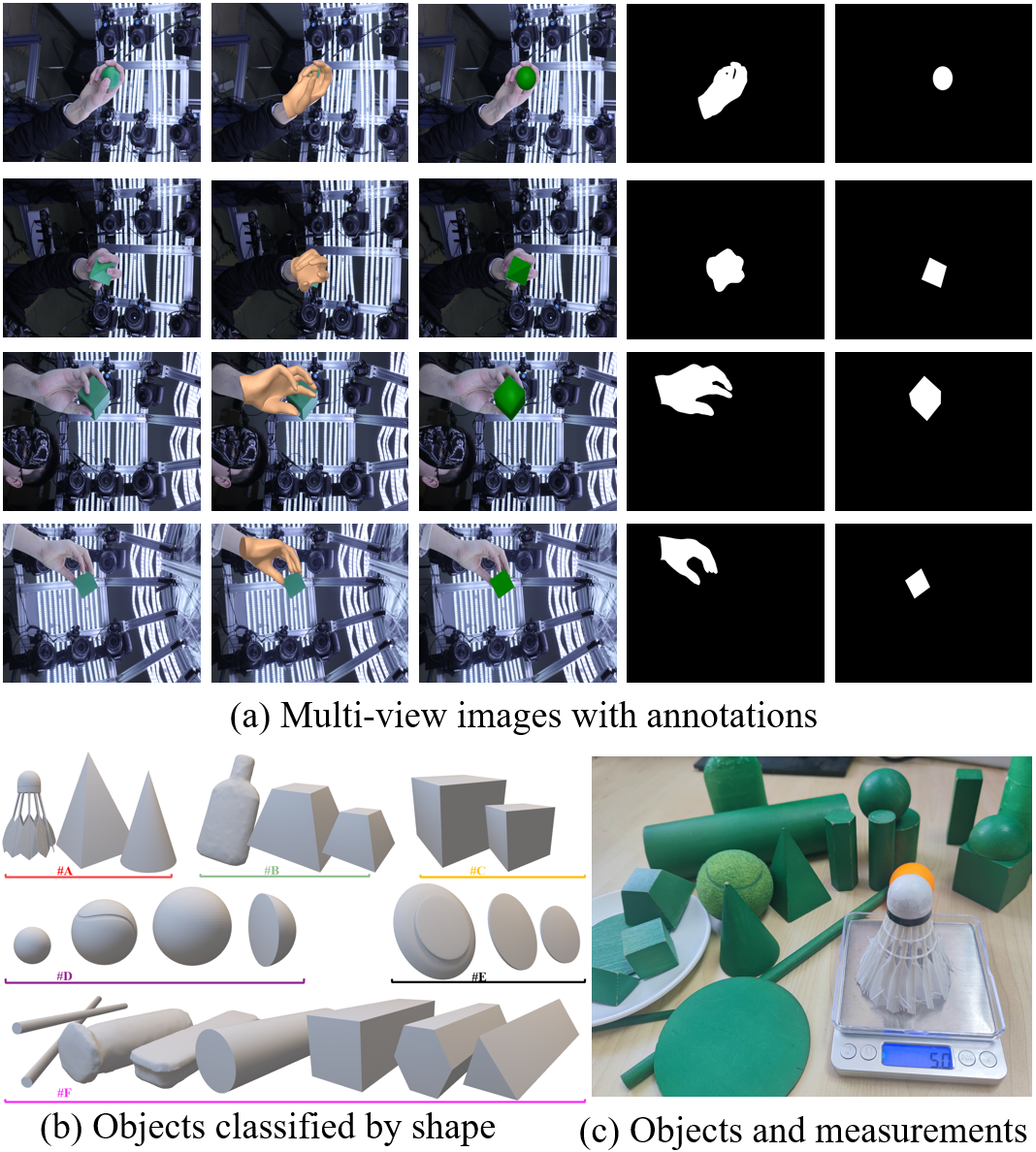}
    \caption{\textbf{Interaction dataset with physical attributes}. (a) Multi-view dataset with mesh and mask annotations; (b) \numobj objects classified into 6 categories; (c) Real models of our objects. }
    \vspace{-6mm}
    \label{fig06_mocapdata}
\end{figure}


\subsection{Interaction Dataset Preparation} 
As shown in~\figmk\ref{fig06_mocapdata}, we created a dataset containing multi-view color images, hand and object visible masks, physical attributes, and stability degree measured by the magnitude of the extra balancing force. In summary, it contains \numscene scenes of \numsbj subjects interacting with \numobj objects captured by \numcam cameras. These objects are classified into 6 categories according to their shape, including A) cones, B) prisms, C) cubes, D) spheres, E) disks, and F) columns. For more details about our dataset, please refer to \supmat. 

\begin{figure*}[!t]
    \centering
    \includegraphics[width=\linewidth]{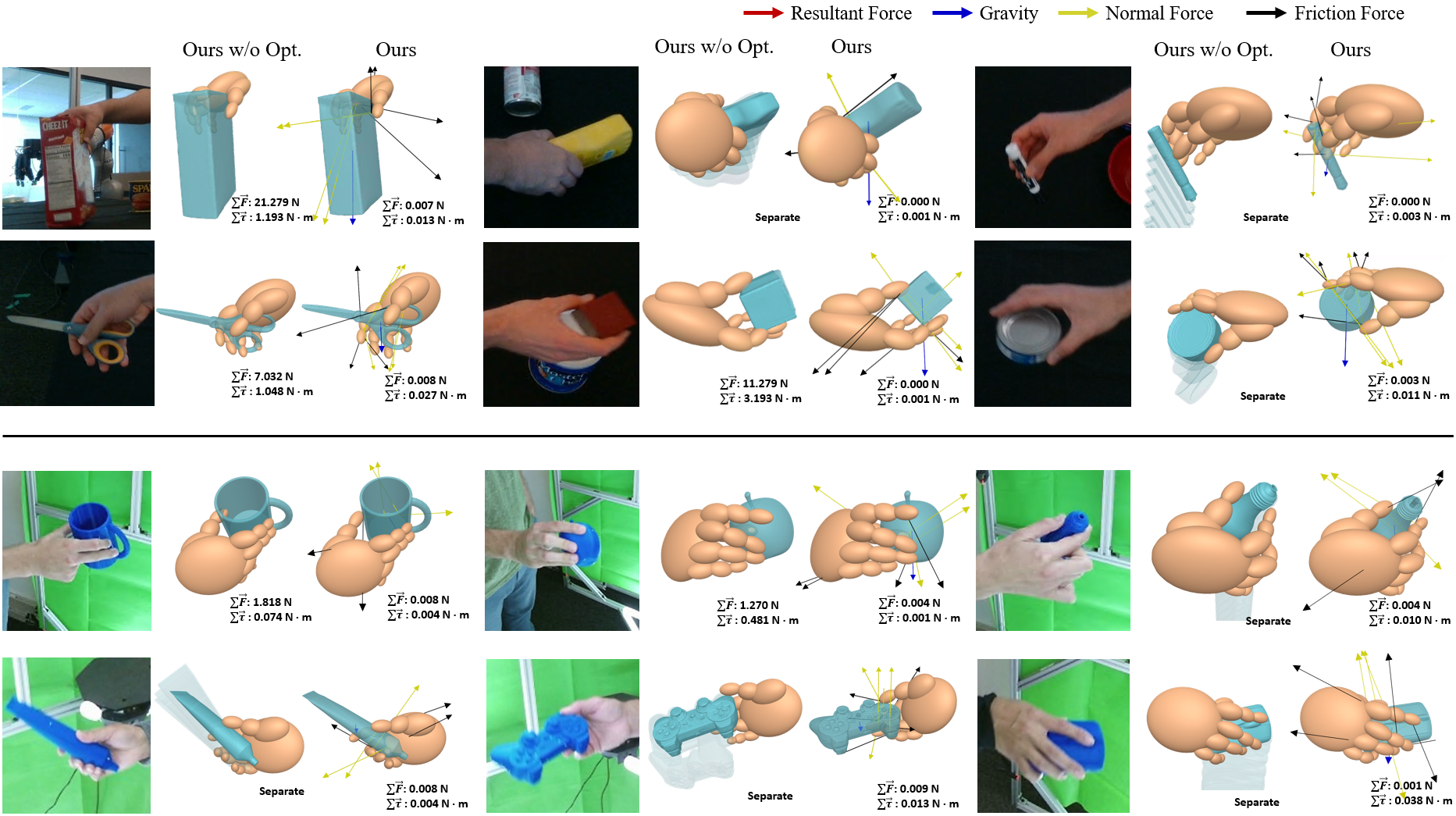}
    \caption{\textbf{Qualitative results on DexYCB~\cite{chao2021dexycb} and ContactPose~\cite{brahmbhatt2020contactpose}}. For each sample, the optimized state increases the stability of the contact while ensuring the consistency of the initial state estimated by the front-end network. }
    \vspace{-2mm}
    \label{fig04_resshow}
\end{figure*}

\section{Experiments} 
In this section, the evaluating datasets and the criteria are first defined in \secmk\ref{sec41_edata}. Our method is compared with the SOTA methods in \secmk\ref{sec43_comp}. Detailed ablation studies are also conducted to our key components in \secmk\ref{sec44_ablation}. 

\subsection{Datasets and Metrics}
\label{sec41_edata}
\noindent\textbf{Datasets}. The existing dataset contains two main types. ~The first type~\cite{garcia2018first, hampali2020honnotate, chao2021dexycb} records real RGB images and the whole hand-object interaction process including approach, contact, and manipulation. This kind of data is used to test our entire pipeline. To reduce the ambiguity in the selection of interacted objects, we follow the method\cite{hasson2019learning, yang2021cpf} to filter these datasets with the 3D distance between the hand-object not exceeding $5 \mathrm{mm}$ as the threshold. The official testing set from HO3D~\cite{hampali2020honnotate} is not used due to the lack of hand mesh ground truth. In the end, the data used for testing contains 7,373 samples in FPHB~\cite{garcia2018first}, 69,292 samples in HO3Dv3~\cite{hampali2020honnotate}, and 93,264 samples in DexYCB~\cite{chao2021dexycb}. Another type~\cite{hasson2019learning,brahmbhatt2020contactpose,taheri2020grab} focuses on recording the contact pattern of the hand-object. For each sequence in GRAB~\cite{taheri2020grab}, we extracted the interaction sub-sequences containing contact between the right hand and object with 50 frames as an interval. This dataset is denoted as GRAB$_{rh50}$. In the end, the data used for testing contains 2,259 samples in ContactPose~\cite{hampali2020honnotate} and 19,008 samples in GRAB$_{rh50}$~\cite{taheri2020grab}. 

\begin{table}[t]
    \rowcolors{1}{}{lightgray}
    \begin{center}
        \resizebox{1\linewidth}{!}{
    \begin{tabular}{c|ccc|ccc}
    \noalign{\hrule height 1.5pt}
    Datasets 
    &\multicolumn{3}{c|}{ContactPose~\cite{brahmbhatt2020contactpose}} 
    &\multicolumn{3}{c}{GRAB$_{rh50}$}\\  
    \midrule
    \rowcolor{white}Methods 
    &GT. &~\cite{grady2021contactopt} &Ours$^\ddagger$ 
    &GT. &~\cite{grady2021contactopt} &Ours$^\ddagger$ 
    \\  
    \midrule
    Max Pene.$(\mathrm{mm})\downarrow$ 
        &11.62 &12.07 &\textbf{8.54}     
        &10.33 &12.38 &\textbf{7.54}  \\   
    Inter.$(\mathrm{cm}^3)\downarrow$
        &12.24 &12.35 &\textbf{6.13}     
        &14.62 &13.97 &\textbf{7.28} \\ 
    Disp. $(\mathrm{mm})\downarrow$ 
        &4.68 &4.35 &\textbf{1.02}     
        &4.25 &4.47 &\textbf{1.23}  \\   
    SC. $\downarrow$
    &1.46 &1.03 &\textbf{0.27}     
        &1.34 &1.28 &\textbf{0.44}  \\   
    \noalign{\hrule height 1.5pt}
    \end{tabular}
    }
    \end{center}
    \vspace{-4mm}
    \caption{\textbf{Evaluations for Hand-Object Contact Estimation. } `Ours$^\ddagger$'' denotes our method with optimization only.}
    \label{tab02_compareContact}
    \vspace{-4mm}
\end{table}

\noindent\textbf{State Error}. Due to the hand mesh reconstructed by our method being different from MANO~\cite{romero2017embodied}, the mean per-point position error (\emph{MPJPE}) of 21 hand joints is chosen to evaluate the 3D reconstruction error. In 2D, the mean intersection over union (\emph{mIOU}) is adopted to evaluate the re-projection error between the conversed mesh and the ground truth. As for the object, the vertices of the posed object are obtained by aligning the object reference mesh with the estimated ellipsoids. The mean per-vertex position error (\emph{MPVPE}) and the \emph{mIOU} are adopted to evaluate the object error. 

\noindent\textbf{Contact Quality}. First, \emph{max penetration} (Max Pene.) and \emph{intersection volume} (Inter.) \cite{hasson2019learning} are adopted to evaluate the geometric relationship. Then, \emph{simulation displacement} (Disp.) \cite{hasson2019learning} and our \emph{stability cost} (SC.) defined in \secmk\ref{sec32_contactopt} are used to evaluate the contact stability in the same simulation settings. For a fair comparison, the ellipsoid hand is converted to the convex-hull mesh when computing these intersection metrics according to \secmk\ref{sec31_hope}. 

\noindent\textbf{Sim-to-Real Gap}. For each scene in our dataset, the correlation between the balancing force and the corresponding cost is used to evaluate the simulation effectiveness.

\subsection{Comparisons}
\label{sec43_comp}
\noindent\textbf{State Estimation}. In the task of estimating hand-object state from monocular images, our method is compared with the methods using pure regression~\cite{hasson2019learning, hasson2020leveraging} and the methods with additional optimization~\cite{yang2021cpf, cao2021reconstructing}. As shown in \tablemk\ref{tab01_compareState}, the hand-object state estimated from our front-end network has a better performance than the direct regressing method, and our full pipeline achieves the best results across data sets. This demonstrates that our approach outperforms other MANO-based regressions in terms of representation, and our recovery module can achieve effective optimization of the contact pattern. In some cases, the position accuracy of hand-object may be slightly influenced by the optimization with their stability increasing. This may be caused by the difference between real and simulated conditions. 

\noindent\textbf{Contact Recovery}. By taking the hand-object state, our recovery module is compared with~\cite{grady2021contactopt} under ContactPose and GRAB$_{rh50}$. As shown in \tablemk\ref{tab02_compareContact}, our method increases the stability of the contact while reducing the penetration. This further illustrates that the contact has been optimized more comprehensively with our method. 

\begin{table*}[!t]
    \rowcolors{1}{}{lightgray}
    \begin{center}
        \resizebox{1\linewidth}{!}{
    \begin{tabular}{c|ccccc|ccccc|ccc}
    \noalign{\hrule height 1.5pt}
    Datasets 
    &\multicolumn{5}{c|}{FPHB~\cite{garcia2018first}} 
    &\multicolumn{5}{c|}{HO3Dv3~\cite{hampali2020honnotate}}
    &\multicolumn{3}{c}{DexYCB~\cite{chao2021dexycb}}\\
    \midrule
    \rowcolor{white}Methods 
    &~\cite{hasson2019learning} &~\cite{hasson2020leveraging} &~\cite{yang2021cpf} &Ours$^\dagger$& Ours 
    &~\cite{hasson2020leveraging} &~\cite{yang2021cpf} &~\cite{cao2021reconstructing} 
    &Ours$^\dagger$ &Ours 
    &GT. &Ours$^\dagger$ &Ours\\ 
    \midrule
    $mIoU_H(\%)\uparrow$   
        &  -   &54.54 &- &59.34 &\textbf{62.01} 
        &64.04 &- &- &\textbf{61.52} &61.43 
        &- &62.64 &\textbf{63.52}\\ 
    $\text{MPJPE}_{H}(\mathrm{mm})\downarrow$
        &28.80 &19.32 &- &19.10 &\textbf{18.56} 
        &14.32 &- &9.50 &10.96 &\textbf{9.14} 
        &- &11.32 &\textbf{11.15}\\ 
    $mIoU_O(\%)\uparrow$ 
        &- &66.10 &- &71.34 &\textbf{72.58} 
        &75.26 &- &- &\textbf{82.53} &{82.47} 
        &- &80.66 &\textbf{81.34}\\
    $\text{MPVPE}_O(\mathrm{mm})\downarrow$  
        &- &21.07 &21.57 &21.14 &\textbf{20.96} 
        &20.08 &73.28$^\diamond$ &- &\textbf{19.34} &19.45 
        &- &18.61 &\textbf{18.84}\\
    Max Pene.$(\mathrm{mm})\downarrow$  
        &15.12 &18.08 &16.92 &15.07 &\textbf{11.43} 
        &10.29 &16.47 &- &16.85 &\textbf{11.36} 
        &10.65 &7.32 &\textbf{6.72}\\
    Inter.$(\mathrm{cm}^3)\downarrow$
        &10.90 &11.05 &11.76 &10.12 &\textbf{6.23} 
        &12.26 &7.44 &- &7.32 &\textbf{6.19} 
        &14.76 &6.94 &\textbf{6.61}\\
    \noalign{\hrule height 1.5pt}
    \end{tabular}
    }
    \end{center}
    \vspace{-4mm}
    \caption{\textbf{Evaluations for Hand-object State Estimation.} ``Ours$^\dagger$'' denotes our method without optimization, ``Ours'' denotes our full pipeline. The item marked by ``-'' indicates that the work has not been trained or tested on the relevant dataset. The item marked by ``$\diamond$'' denotes the wrist-relative object vertex error. }
    \label{tab01_compareState}
    \vspace{1mm}
\end{table*}

\subsection{Ablation Study}
\label{sec44_ablation}
Due to the ContactPose~\cite{brahmbhatt2020contactpose} having both images and accurate contact, most of our ablation experiments are based on this dataset. Among them, the results of completely using our entire pipeline are in the last row of \tablemk\ref{tab03_ablationContact}.

\noindent\textbf{Training Paradigms}. The verifications of two key components in the training process, including semi-supervised pre-training and contact loss, are shown in the first two rows of \tablemk\ref{tab03_ablationContact}. The lack of collision loss may worsen the initial state of the hand-object before optimization, which in turn affects the whole optimization process. On the other hand, the network without pre-training is less robust to the diversity of hand posture, the change of perspectives, and the occlusion during hand-object interaction, which may have similar effects on the entire pipeline. 

\noindent\textbf{Stability Cost}. We compared the importance of each term in our stability cost, as shown in the middle 6 rows of \tablemk\ref{tab03_ablationContact}. The lack of each item would weaken the final result, among which the force item has the greatest impact. Further, we replaced the stability cost with the displacement defined on our hand model as the objective of driving optimization, while the result becomes worse as shown in row 8 of \tablemk\ref{tab03_ablationContact}. The main reason may be that the object displacement can only reflect the contact stability in fewer simulation steps. Therefore, our criteria could measure contact patterns more generally. The method with only collision detection in the physics engine is also employed, which does not have enough stability either. 

\noindent\textbf{Hand Model}. As shown in row 9 and row 10 of \tablemk\ref{tab03_ablationContact}, the choices of collision shape and local DoFs of our hand model are also explored under the same simulation conditions. Among them, the hand consisting of mesh segments leads to poorer stability. This may be caused by the fact that the mesh collision shape in the physics engine is automatically approximated as a convex hull, which changes the accuracy of collision detection. On the other hand, the hand with more local DoFs has lower accuracy because it increases the difficulty of optimization. To improve the efficiency of sampling and optimization methods, the methods with 20 local DoFs were adopted. 
 
\begin{figure}[!t]
    \centering
    \includegraphics[width=\linewidth]{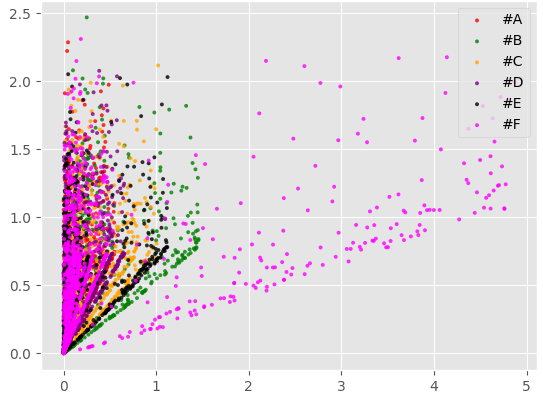}
    \caption{\textbf{Correlation between stability cost and real forces}. The horizontal axis is the real-measured force, and the vertical axis is the corresponding stability cost of the same hand-object state reconstructed in the simulation environment. The data corresponding to 6 types of objects are marked with different colors. }
    \vspace{-4mm}
    \label{fig07_correlation}
\end{figure}

\noindent\textbf{Sim-to-Real Correlation}. With interaction scenes in our dataset, we quantitatively analyze the relationship between simulation stability cost and actual compensation force. For more details about the measurements of the balancing forces and corresponding physical properties, please refer to our \supmat. In the experiment, each hand-object scene reconstructed was used to directly initialize our simulation interaction scene, and then their stability cost was calculated. Each reconstructed hand object scene is used to directly initialize our simulation interaction scene. Their actual stability during capturing is measured by the scale of the balancing force, and the stability in the simulation is measured by the stability cost. The mass and friction coefficient of the object in the simulation is set to be the same as the actual measured values. As shown in \figmk\ref{fig07_correlation}, for objects with different shapes, the actual stability and simulation stability have different correlations. Among them, objects in class F (\thatis columns) correspond to multiple slopes, which is caused by the great scale variations within the categories.

\begin{table}
    \rowcolors{1}{}{lightgray}
    \begin{center}
        \resizebox{1\linewidth}{!}{
          \begin{tabular}{c| ccc}
                \noalign{\hrule height 1.5pt}
                Method & Inter.$(\mathrm{cm}^3)\downarrow$ & Disp. $(\mathrm{mm})\downarrow$ & SC. $\downarrow$\\
                \midrule
                 w/o ${L}_{in}$ & 7.41 & 4.65 &0.86\\
                 w/o pre-trained & 7.34 & 3.77 & 0.51\\
                \midrule
                 w/o $C_{\mathrm{S}}$ Opt. & 6.32 & 3.43 & 4.62\\
                 w/o $L_{cnt}$. & 6.28 & 2.39 & 0.58\\
                 w/o $L_{frc}$. & 6.23 & 2.66 & 0.73\\
                 w/o $L_{tau}$. & 6.37 & 2.17 & 0.64\\
                 w/o $C_{stab}$ Opt. & 7.32 & 1.92 & 1.44\\
                 Opt. with Disp. & 6.94 & 3.43 & 4.62\\
                \midrule
                 w/o Ellipsoids & 6.36 & 1.59 & 0.64\\
                 with $|\bm{\vartheta}|= {48}$ &6.32 & 1.47 & 0.47\\
                \midrule
                Ours &  \textbf{6.24} & \textbf{1.13} & \textbf{0.31}\\
                \noalign{\hrule height 1.5pt}
          \end{tabular}
        }
    \end{center}
    \caption{\textbf{Ablation study on ContactPose.} 
    The components in network training paradigm, optimization function and physical hand model are evaluated. }
    \label{tab03_ablationContact}
\end{table}
\section{Conclusion} 
This paper proposes a novel monocular hand-object contact recovery scheme driven by the simulated stability criteria in the physics engine. Through sampling-based optimization, a more stable contact pattern is obtained without data prior dependence. A hand-object ellipsoid representation further promotes the effective implementation of our regression-optimization pipeline. It enables personalized hand shape variations at the same time. The sim-to-real consistency is verified later by our contact scene dataset with real physical properties and stability evaluation. 

\noindent\textbf{Limitations and Future Work}. Although our method is robust under existing datasets, it may become invalid in a complex scene with severe occlusion or multiple hands/objects. Getting rid of object mesh dependence is also significant for the improvement of our approach. In the future, rewards with our stability cost considerations could more effectively guide reinforcement learning methods to reconstruct hand-object interaction sequences.
\clearpage
\begingroup

\twocolumn[
\begin{center}
    {\Large \bf \Large{Stability-driven Contact Reconstruction From Monocular Color Images} \\ -- Supplementary Material -- \par}
  \vspace*{30pt}
\end{center}
]
\appendix

\setcounter{table}{0}
\setcounter{figure}{0}
\setcounter{equation}{0}
\renewcommand{\thetable}{\thesection.\arabic{table}}
\renewcommand{\thefigure}{\thesection.\arabic{figure}}
\renewcommand{\theequation}{\thesection.\arabic{equation}}

\section{Overview}
In this supplementary document, we first introduce our interaction dataset, named \VCLhoDBname (Contact with Balancing Force), which contains physical attributes of hand-object and stability degree measured by the magnitude of the extra balancing force~(\secmk\ref{sec_System}). More ablation studies of our sampling-based optimization are discussed in \secmk\ref{sec_Alternatives}. They were not included in the paper due to the page limit.


\begin{figure}[!t]
    \centering
    \includegraphics[width=\linewidth]{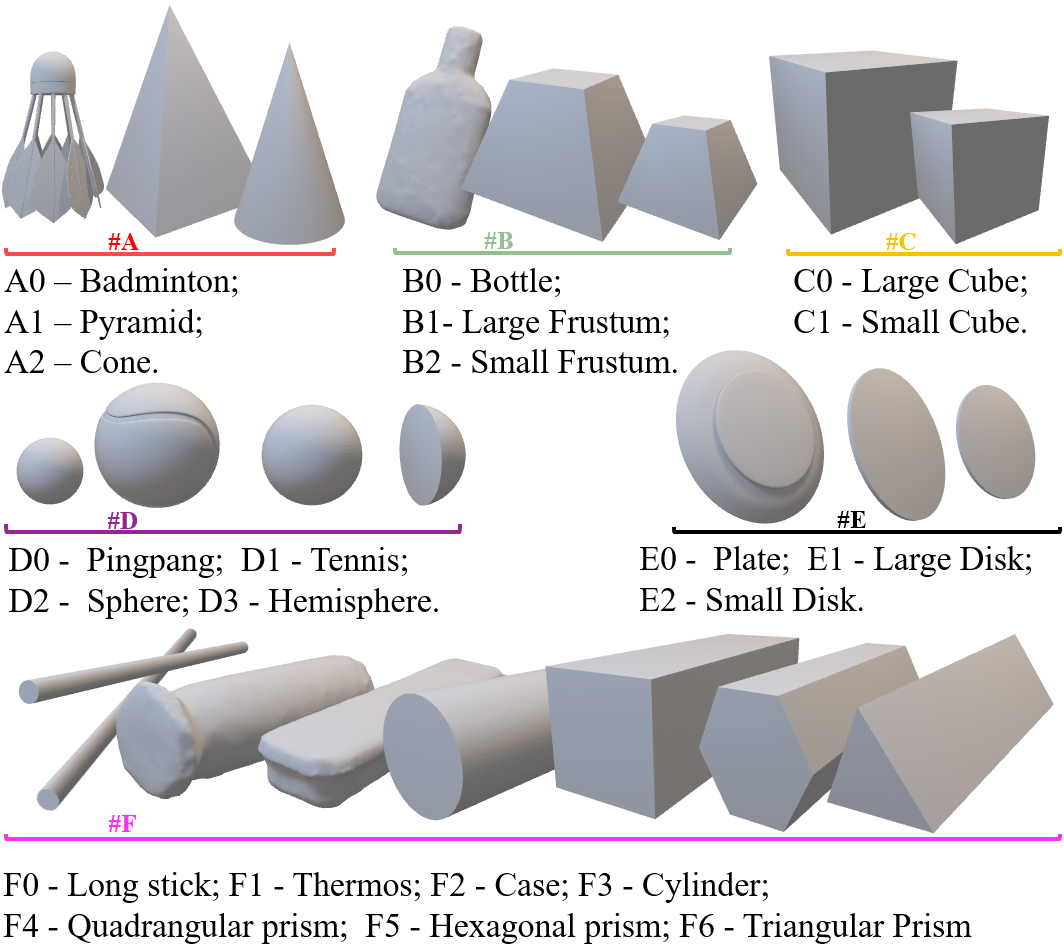}
    \caption{\textbf{Objects and corresponding IDs in each category}. There are \numobj objects classified into 6 categories according to their shape, and each category contains multiple regular geometries and a multi-view reconstruction model.}
    \vspace{-4mm}
    \label{fig02_objectID}
\end{figure}

\begin{figure*}[!t]
    \centering
    \includegraphics[width=\linewidth]{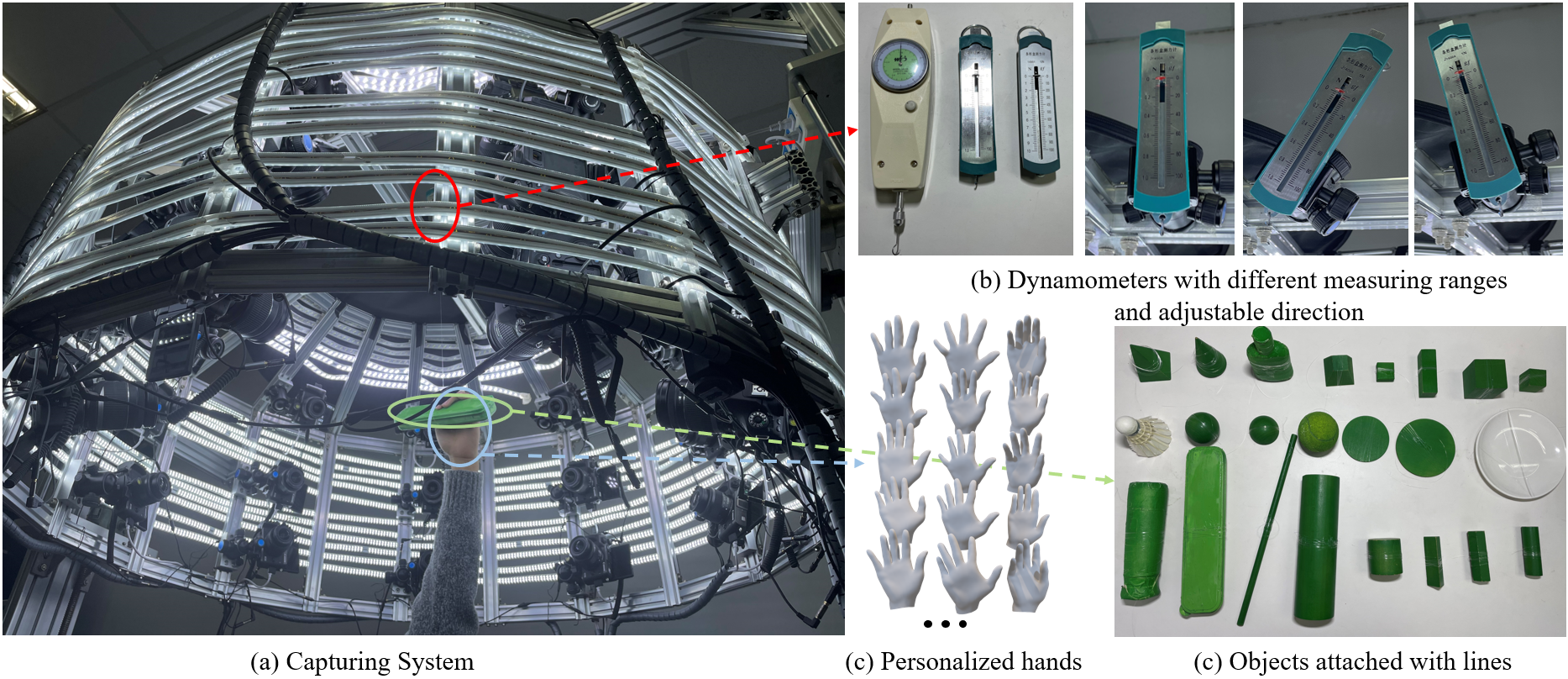}
    \caption{\textbf{Capturing System}. Multi-view system with balancing force measuring devices.}
    \vspace{-4mm}
    \label{fig01_system}
\end{figure*}
\section{Capturing System}
\label{sec_System}
The experimental image data was captured by \numcam calibrated cameras capable of capturing images. The capturing signal of the camera system is triggered when the indication of the dynamometer and the hand object interaction state is no longer changed. 
During the capture process, the multi-view system records the 3D state of the hand object, and the hand-object contact includes the stable states and the unstable states. As shown in \figmk\ref{fig01_system}, the additional external force needed to maintain the object balance is recorded by a dynamometer with a suitable range suspended from the top of the system. 

\begin{table}   
    \rowcolors{1}{}{lightgray}
    \begin{center}
    \resizebox{0.95\linewidth}{!}{
    \begin{tabular}{c c c c}
    \noalign{\hrule height 1.5pt}
    Objects (ID) & Size (mm) & Mass (g) & Friction  \\
    \midrule
    A0 & $(67, 67, 72)$ &  5.1 & 0.010 \\ 
    A1 & $(49, 49, 77)$ & 46.7 & 0.659 \\ 
    A2 & $(50, 50, 77)$ & 0.636 & 0.673 \\ 
    \midrule
    B0 & $(64, 28, 131)$& 123.9 & 0.641 \\ 
    B1 & $(48, 48, 77)$ & 37.5 & 0.647 \\ 
    B2 & $(30, 30, 48)$ & 18.3 & 0.621 \\ 
    \midrule
    C0 & $(50, 50, 50)$ & 87.1 & 0.689 \\ 
    C1 & $(29, 29, 29)$ & 18.9 & 0.423 \\ 
    \midrule
    D0 & $(40, 40, 40)$ & 3.0 & 0.086 \\ 
    D1 & $(66, 66, 66)$ & 56.2 & 0.854 \\ 
    D2 & $(48, 48, 48)$ & 44.8 & 0.670 \\ 
    D3 & $(48, 48, 24)$ & 23.4 & 0.513 \\ 
    \midrule
    E0 & $(152, 152, 20)$ & 95.8 & 0.412 \\ 
    E1 & $(100, 100, 3)$ & 13.7 & 0.342 \\ 
    E2 &  $(80, 80, 2)$ & 6.4 & 0.336 \\ 
    \midrule
    F0 & $(10, 10, 300)$ & 10.9 & 0.381 \\ 
    F1 & $(50, 50, 162)$ & 39.8 & 0.653 \\ 
    F2 & $(30, 58, 245)$ & 67.8 & 0.443 \\ 
    F3 & $(23, 23, 75)$ & 414.2 & 0.687 \\ 
    F4 & $(25, 25, 76)$ & 37.1 & 0.691 \\ 
    F5 & $(29, 29, 75)$ & 27.2 & 0.748 \\ 
    F6 & $(23, 23, 75)$ & 15.9 & 0.722 \\ 
    \noalign{\hrule height 1.5pt}
    \end{tabular}
    }    
    \end{center}
    \caption{\textbf{Object physical attributes.} Our dataset provides physical attributes for each object, including object size, mass, and friction coefficient between the object and the hand (average from multiple measurements).} 
    \label{tab_PhyPara}
    \vspace{-4mm}
\end{table}

\noindent\textbf{Objects}. 
As shown in \figmk\ref{fig02_objectID}, there are \numobj objects classified into 6 categories according to their shape: cones, prisms, cubes, spheres, disks, and columns. On average, each category contains multiple regular geometries and one everyday object. Among them, the regular geometries are all made of the same wood. Most objects are painted green to facilitate segmentation from the image. The object meshes were acquired in three ways: 
\begin{itemize}
    \item Scan reconstruction. The meshes of B0, F1, F2 were created using the multi-view reconstruction method. In this step, an additional texture~\cite{li2013multiple} is applied to the surface of the object; 
    \vspace{-2mm}
    \item CAD modeling. The meshes of the regular geometries were created in the modeling software based on the measured sizes;
    \vspace{-2mm}
    \item Online Searching. The meshes of A0, D0, D1, and E0 were first downloaded from Thingiverse~\cite{Thingiverse2008}. After that, the size of each mesh was fine-tuned according to the size of our real object. 
\end{itemize}
The physical attributes of all interacted objects are recorded in \tablemk\ref{tab_PhyPara}. The size of the object corresponds to the length, width, and height of the object mesh bounding box in the rest pose. The mass of each object was weighed by a high-precision digital electronic scale. The friction between each object and flat subject skin was measured by the dynamometer in a dry environment. Each physical quantity was measured multiple times and averaged to minimize error. When preparing training
data, the radii of the object ellipsoid correspond to the size of
the object bounding box in its rest pose. The 6 DoF pose
of the ellipsoid is consistent with the object transformation
from the rest pose to the pose in images, which is annotated in most existing datasets. Although similar radii or symmetry
in an object mesh may lead to confusion into those annotations, this risk seems to be common to the pipeline
represented object pose with an oriented bounding box.

\noindent\textbf{Subjects}. A total of \numsbj people were invited to participate in the production of the dataset, including 10 males and 10 females. All the involved subjects agreed to the release of the dataset, and their consent forms are on the last page of this supplementary. The friction coefficient was remeasured for each subject, and column 4 of \tablemk\ref{tab_PhyPara} reflects the average value. As shown in \figmk\ref{fig01_system}(c), the personalized hand parameters of the subject are pre-optimized by the multi-view system. In the subsequent hand-object state reconstruction, only the pose parameters of the personalized hand are optimized. 

\begin{figure*}[!t]
    \centering
    \includegraphics[width=\linewidth]{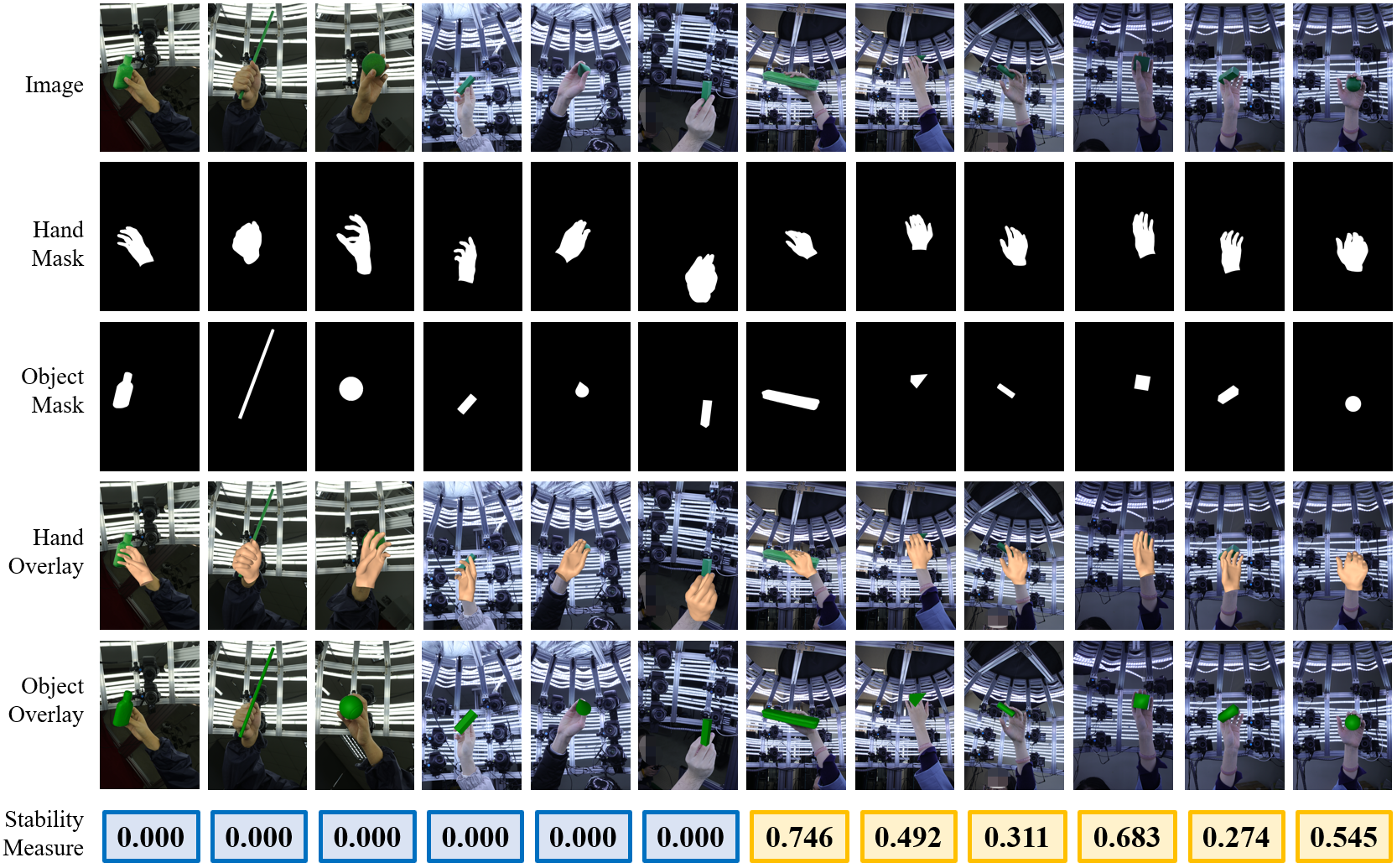}
    \caption{\textbf{Interaction dataset with balancing force measurements}. Each column corresponds to an interaction scene. The first 6 columns show the scene without threads, and the last 6 columns show the scene with the thread and balancing force on the thread. To facilitate the display, only one of the \numcam perspective images is selected for display. }
    \vspace{-2mm}
    \label{fig03_CBF}
\end{figure*}

\noindent\textbf{Markless Capture}. 
Each hand-object interaction state is observed through \numcam calibrated cameras. Most of these camera models are Canon EOS M6 with high resolution. An additional camera facing the dynamometer was used to record the magnitude of the equilibrium force.  All these cameras were connected by the common audio cable for synchronization. The trigger signal was sent out of the system by a high-speed wireless remote controller. For each RGB frame, we performed instance segmentation to get hand mask and object mask manually. Then we ran the SRHandNet~\cite{wang2019srhandnet} to locate the key-points $\bm{x}^\star \in \mathbb{R}^{2\times21}$. Using the relative position relationship between the object and the hand, the key-point order on the object is defined and marked in each perspective. By minimizing the re-projection error, this 2D information from multi-view images is used to optimize the shape and pose for the 3D object mesh and the hand LBS template. 

\noindent\textbf{Force Application}. 
As shown in \figmk\ref{fig01_system} (b), a dynamometer is hoisted on the top of the system. To flexibly measure forces from different directions, it is mounted on a tripod head with 3 DoF rotation. Due to the large difference in mass of all our objects, the dynamometers have different ranges and accuracies: (1) The range is $20$N and the resolution is $0.1$N; (2) The range is $10$N and the resolution is $0.05$N; (3) The range is $1$N and the resolution is $0.02$N. We chose thin threads made of UHMWPE (Ultra-high-molecular-weight polyethylene) material to connect the dynamometer and the interacted object. It is transparent and only $0.14 \mathrm{mm}$ in diameter. Some originally stable scenes without thin balancing thread are also captured. As shown in the first six columns and the last six columns of \figmk\ref{fig03_CBF}, this thin thread does not significantly affect the appearances of the interacted images. The capturing signal of the camera system was triggered when the indication of the dynamometer and the hand object interaction state is no longer changed. For each subject contacting each object, 5 actions with obvious differences in stability (indicated by the dynamometer) are collected. To facilitate postural fixation, the elbow participant was supported by the table plate during the capturing. To reduce the error, each scene is repeated 5 times through the above process, and the mean magnitude of the force is finally recorded. 


\section{More Alternatives}
\label{sec_Alternatives}
In this section, we provide evaluations of more variants with the same evaluating conditions as our main paper. These experiments focus on comparing the effects of different hyper-parameters in our sampling process on the final results. Among them, the number of samples $N$ has the greatest impact on the results. Although a larger number of samples gives better results, for the hand model with 23 DoFs, we find that setting it to 300 is sufficient, and the improvement in results from more samples does not compensate for the longer running time of the algorithm. The number of sampling iterations $K$ has a relatively small effect on the results. The number of the simulation time step $T$ has a truncated effect on the method. When the step amount exceeds 150 steps, the results hardly change. ~ We find that the
final stable contact pattern obtained by our method is hardly
affected when the physical parameters are varied within the same order of magnitude. Nevertheless, considering the simulation stability~\cite{coumans2021}, the recommended values should be close to the parameters measured in our dataset.

\begin{table}
    \rowcolors{1}{}{lightgray}
    \begin{center}
        \resizebox{1\linewidth}{!}{
          \begin{tabular}{c| ccc}
                \noalign{\hrule height 1.5pt}
                Method & Inter.$(\mathrm{cm}^3)\downarrow$ & Disp. $(\mathrm{mm})\downarrow$ & SC. $\downarrow$\\
                \midrule
                $K = 20$ & 6.28 & 1.15 & 0.36\\
                $K = 40$ & 6.23 & 1.12 & 0.31\\

                $N = 100$ & 7.03 & 1.13 & 0.39\\
                $N = 500$ & 6.12 & 1.09 & 0.26\\

                $T = 60$ & 6.35 & 1.31 & 0.41\\
                $T = 240$ & 6.24 & 1.13 & 0.31\\
                \midrule
                Ours &  \textbf{6.24} & \textbf{1.13} & \textbf{0.31}\\
                \noalign{\hrule height 1.5pt}
          \end{tabular}
        }
    \end{center}
    \vspace{-4mm}
    \caption{\textbf{More Ablation study on ContactPose~\cite{brahmbhatt2020contactpose}.} 
    The components in network training paradigm, optimization function and physical hand model are evaluated. }
    \vspace{-4mm}
    \label{tab03_ablationContact}
\end{table}

\begin{figure*}[!t]
    \centering
    \includegraphics[width=\linewidth]{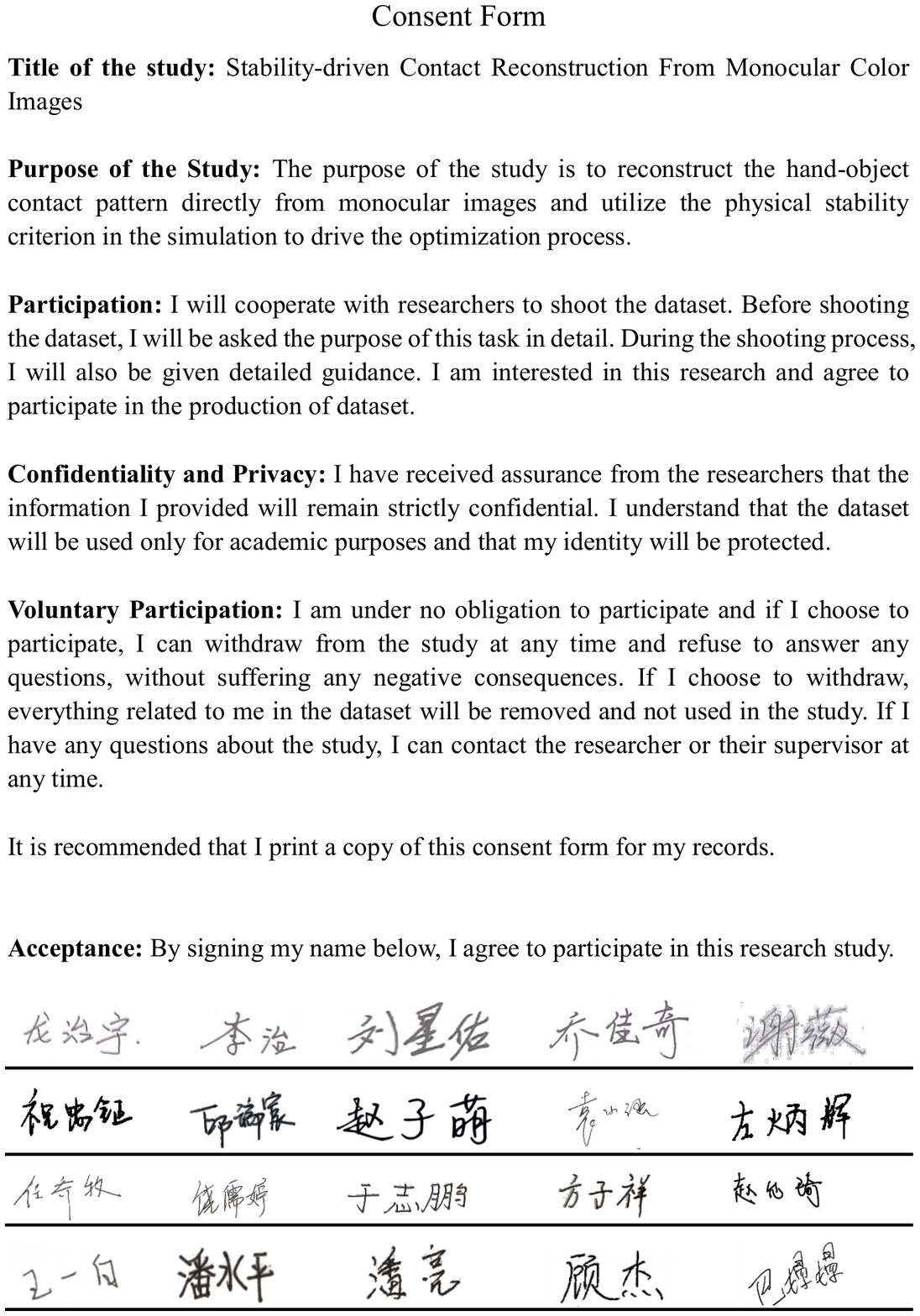}
    \vspace{-4mm}
    \label{fig04_content}
\end{figure*}

{\small
\bibliographystyle{ieee_fullname}
\bibliography{egbib}
}

\end{document}